\documentclass{article}

\usepackage[preprint,nonatbib]{neurips_2020}

\usepackage[utf8]{inputenc} 
\usepackage[T1]{fontenc}    
\usepackage{hyperref}       
\usepackage{url}            
\usepackage{booktabs}       
\usepackage{amsfonts}       
\usepackage{nicefrac}       
\usepackage{microtype}      

\usepackage{amsmath,amssymb,amsthm}
\usepackage{mathtools}

\usepackage{algorithm}
\usepackage{algorithmic}
\usepackage{amsmath}

\usepackage{graphicx}
\usepackage{caption}
\usepackage{subfigure}

\def\tg{\tilde{g}}
\def\tm{\tilde{m}}
\def\tv{\tilde{v}}

\def\R{{\mathbb R}}
\def\CE{\mathbf{E}}
\def\mG{\mathcal{G}}

\graphicspath{{images/}} 

\title{Accelerated Large Batch Optimization of BERT Pretraining in 54 minutes}

\author{
  Shuai Zheng, Haibin Lin, Sheng Zha, Mu Li\\
  \{{shzheng, haibilin, zhasheng, mli\}@amazon.com} \\
  Amazon Web Services 
}

\begin{document}

\maketitle

\begin{abstract}
 BERT has recently attracted a lot of attention in natural language understanding (NLU) and achieved state-of-the-art results in various NLU tasks. However, its success requires large deep neural networks and huge amount of data, which result in long training time and impede development progress. Using stochastic gradient methods with large mini-batch has been advocated as an efficient tool to reduce the training time. Along this line of research, LAMB is a prominent example that reduces the training time of BERT from 3 days to 76 minutes on a TPUv3 Pod. In this paper, we propose an accelerated gradient method called LANS to improve the efficiency of using large mini-batches for training. As the learning rate is theoretically upper bounded by the inverse of the Lipschitz constant of the function, one cannot always reduce the number of optimization iterations by selecting a larger learning rate. In order to use larger mini-batch size without accuracy loss, we develop a new learning rate scheduler that overcomes the difficulty of using large learning rate. Using the proposed LANS method and the learning rate scheme, we scaled up the mini-batch sizes to 96K and 33K in phases 1 and 2 of BERT pretraining, respectively. It takes 54 minutes on 192 AWS EC2 P3dn.24xlarge instances to achieve a target F1 score of 90.5 or higher on SQuAD v1.1, achieving the fastest BERT training time in the cloud. A fast implementation of LANS is available online \footnote{\url{https://github.com/szhengac/apex/blob/lans/apex/optimizers/fused_lans.py}}. 
\end{abstract}

\section{Introduction}

Deep neural networks
have achieved remarkable performance in various tasks such as image classification \cite{he2016deep}, speech recognition \cite{graves2013speech}, machine translation \cite{vaswani2017attention}, and natural language understanding \cite{devlin2019bert}.  
These problems are typically formulated as the minimization of a nonconvex objective on a set of training samples. 
The most popular optimization tool is stochastic gradient descent (SGD)  \cite{robbins1951stochastic, ghadimi2013stochastic, bottou2018optimization}, which is simple and computationally efficient. However, 
deep learning thrives with large model size and huge amount of data, and it raises significant challenge even for a cheap optimizer such as SGD to reach a decent solution in a reasonable amount of time. 
For example, training a large BERT model requires 3 days on 16 TPUs \cite{devlin2019bert} and it takes 40 days to train an AlphaGo Zero system \cite{silver2017mastering}. 
Thus, it is necessary to develop fast optimization methods to accelerate deep neural network training.  

To improve the training efficiency, ones often have to 
distribute the computation of a large mini-batch gradient to multiple computing nodes \cite{dean-12,zinkevich2010parallelized}.  
Distributed synchronous SGD has become a de-facto method for large-scale machine learning problems. For further acceleration, variants that use classic momentum \cite{polyak1964some} and Nesterov's momentum \cite{nesterov1983method} have been widely adopted \cite{sutskever2013importance, he2016deep}. 
By increasing the mini-batch size, distributed synchronous SGD can make use of a larger learning rate so that the total number of training iteration can be reduced accordingly. 
The learning rate typically grows with the square root of the mini-batch size \cite{dekel2012optimal} or can even increase linearly with the mini-batch size when appropriate warmup schedule is employed \cite{goyal2017accurate}.
However, one cannot increase the learning rate indefinitely and the learning rate scaling heuristics that depends on the
mini-batch size can break for some cases \cite{shallue2019measuring}. Thus, it takes more efforts to search for good hyper-parameters for synchronous SGD methods . 

To accelerate the convergence of SGD and spend less work in tuning hyper-parameters, many 
coordinate-wise adaptive learning rate based methods have been introduced \cite{almeida1999parameter,duchi2011adaptive,kingma2014adam,hinton2012rmsprop,zeiler2012adadelta,reddi2018convergence}. Adaptive gradient methods dynamically adjust their  learning rates according to the received noisy gradients. 
On the other hand, several attempts have been made to use layer-wise learning rates for different layers \cite{singh2015layer,yang2017lars,adam2017normalized,zhou2018adashift}. It has been shown that layer-wise learning rate improves generalization performance in practice. Very recently, LAMB is proposed in \cite{you2020reducing}. It combines AdamW optimizer \cite{loshchilov2017decoupled} with normalized gradient descent \cite{nesterov1984,hazan2015beyond}. It is shown that LAMB managed to train BERT with a large mini-batch size of 64K without losing accuracy. However, it cannot further scale up to an even larger mini-batch size. 

In this paper, 
we introduce per-block gradient normalization to LAMB and modify its momentum term by taking advantage of the connection between the classic momentum and Nesterov's momentum.  
The resultant accelerated gradient method is called LANS. 
Moreover, as the linear scaling only works up to certain mini-batch sizes, 
we propose to add a constant learning rate stage after the warmup phase. 
Such change allows the optimizer to use the maximum learning rate for a longer period of time, which results in sufficient training progress even we cannot further increase the maximum learning rate. 
The experimental results show that the proposed methods can use a very large mini-batch size of 96K and reduce the BERT pretraining time to 54 minutes on 192 Amazon EC2 P3dn.24xlarge instances without suffering from any performance deterioration.

{\bf Notations}. 
For a vector $x \in \R^d$, $\sqrt{x}$ is the
element-wise square root of $x$, $x^2$ is the coordinate-wise square of $x$, $\|x\|_2 = \sqrt{x^Tx}$. 
For two vectors $x$ and $y$, 
$x/y$ denotes the element-wise division.

\section{Related Works}

In machine learning, one is interested in minimizing a $\ell_2$-norm regularized optimization problem of the form
\begin{eqnarray} \label{eq:problem}
\min_{x \in \R^d} F(x) = \CE_{\xi}[f(x, \xi)] + \frac{\lambda}{2}\|x\|_2^2,
\end{eqnarray}
where $f$ is some 
possibly nonconvex
loss function, $\xi$ is a random sample, $x$ is the model parameter, $\lambda$ is the regularization parameter, 
and the
expectation is taken w.r.t. the underlying sample distribution. The objective (\ref{eq:problem}) reduces to the expected risk that 
measures the generalization performance
on unseen data \cite{bottou2018optimization} when $\lambda=0$, and
reduces to the regularized empirical risk when a finite training set is considered. 

\subsection{LAMB}

For the gradient $g_t \in \R^d$, let $g_t = [g_{t, \mathcal{G}_1}, g_{t, \mathcal{G}_2}, \dots, g_{t, \mathcal{G}_B}]$ be its decomposition into $B$ blocks, 
where $\mathcal{G}_b$ is the set of indices in block $b$, and $g_{t, \mathcal{G}_b}$ 
is the corresponding block of variables. A block can be a parameter tensor/matrix/vector. 
Recently, You {\em et al.\/} 
\cite{you2020reducing} introduced a layer-wise adaptive large batch optimization method, called LAMB (Algorithm~\ref{alg:lamb}). 
LAMB proposed to add a normalization factor to the AdamW (ADAM with weight decay) \cite{loshchilov2017decoupled} update that divides the update by its $\ell_2$ norm. This ensures that the update
for each block has unit $\ell_2$-norm. And, the learning rate is rescaled by $\phi(\|x_{t, \mG_b}\|_2)$ for some function $\phi: \R^{+} \rightarrow \R^{+}$. In practice, it is generally set to an identity mapping. In this case, the update preserves the same $\ell_2$ norm as the model parameters, and the model parameters change in a smooth trajectory.

\begin{algorithm}[ht]
\caption{LAMB \cite{you2020reducing}}
\label{alg:lamb}
\begin{algorithmic}[1]
    \STATE {\bfseries Input:} step size sequence $\eta_t$; $0 < \beta_1, \beta_2 < 1$; scaling function $\phi$; $\epsilon > 0$; regularization parameter $\lambda$.
    \STATE {\bfseries Initialize:} $x_1 \in \mathbb{R}^d$; $m_0, v_0 = 0$.
    \FOR{$t = 1, \ldots, T$}
        \STATE Compute mini-batch stochastic gradient $g_{t}$
        \FOR{$b=1, 2, \dots, B$}
        \STATE{$m_{t, \mG_b} = \beta_1m_{t-1, \mG_b} + (1 - \beta_1)g_{t, \mG_b}$}
        \STATE{$v_{t, \mG_b} = \beta_2v_{t-1, \mG_b} + (1 - \beta_2)g_{t, \mG_b}^2$}
        \STATE{$\tm_{t, \mG_b} = m_{t, \mG_b} / (1 - \beta_1^t)$}
        \STATE{$\tv_{t, \mG_b} = v_{t, \mG_b} / (1 - \beta_2^t)$}
        \STATE{compute ratio $r_{t, \mG_b} = \frac{\tm_{t, \mG_b}}{\sqrt{\tv_{t, \mG_b}} + \epsilon}$}
        \STATE{$x_{t + 1, \mG_b} = x_{t, \mG_b} - \eta_t \frac{\phi(\|x_{t, \mG_b}\|_2)}{\|r_{t, \mG_b} + \lambda x_{t, \mG_b}\|}(r_{t, \mG_b} + \lambda x_{t, \mG_b})$}
     \ENDFOR
    \ENDFOR
\end{algorithmic}
\end{algorithm}

It was shown in \cite{you2020reducing} that LAMB enables a very large mini-batch size of 64K for training BERT while being able to achieve comparable accuracy to the small mini-batch size. 
With such large mini-batch size, the training time of BERT pretraining reduced from 3 days to 76 minutes on 1024 TPUs.

\subsection{Nesterov Momentum}

Momentum methods have been widely used in training deep networks 
\cite{sutskever2013importance}. 
The classic momentum method, also known as heavy-ball method, introduced in
\cite{polyak1964some} accumulates the past gradients $g_t$'s into a momentum vector $m_t$ (with
$m_{0} = 0$), which serves as a smoothing of the velocity:
\begin{eqnarray} 
m_{t} & = & \mu m_{t-1} + g_{t}  \\
 x_{t+1} & = & x_t -  \eta_tm_t, \label{eq:momentum}
\end{eqnarray}
where $\mu \in [0, 1)$ is the momentum parameter.
For a twice differentiable strongly 
convex function, it is known that the classic momentum method can be used to accelerate the gradient
descent and improve the convergence rate from $O((1 - \kappa)^t)$ to $O((1 - \sqrt{\kappa})^t)$.
where $\kappa$ is the condition number of the functions. 
For training deep neural network, the momentum method accelerates early optimization and helps gradient descent method escape from the local minimums.

Nesterov's accelerated gradient (NAG) 
\cite{nesterov1983method, sutskever2013importance} is another kind of momentum method that is closely
related to the classic momentum method in that it can be written as:  
\begin{eqnarray*} \label{eq:nes-momentum}
m_{t} & = & \mu m_{t-1} + g_{t} \\
 x_{t+1} & = & x_t -  \eta_t(\mu m_{t} + g_{t}). 
\end{eqnarray*}
Expanding (\ref{eq:momentum}) to $x_{t+1}  =  x_t -  \eta_t(\mu m_{t-1} + g_t)$, we can see that we get NAG by replacing $m_{t-1}$ with $m_t$. 
Thus, Nesterov's momentum differs from the classic momentum in that it updates the model parameter using the future momentum vector. One can interpret Nesterov's momentum as an attempt to add a correction direction to the classic momentum method. 
NAG is argued to be more effective in the early optimization, and is more tolerant
of large values of $\mu$ compared to the classic momentum method \cite{sutskever2013importance}.  Recently, Adam with Nesterov's momentum is proposed in \cite{dozat2016incorporating} and it shows better convergence performance than Adam on some tasks. Inspired by this change, two variants of LAMB using Nesterov's momentum are proposed in \cite{you2020reducing}. However, their modifications do not take the normalization factor into account, and the resultant algorithms do not show any improvement over LAMB. In this paper, we propose a different way to modify the momentum component of LAMB to take advantage of the superior performance of Nesterov's acceleration.

\section{Proposed Methods}

\subsection{Normalized Gradient}

In LAMB, the update is normalized by its $\ell_2$ norm. In addition to that, we propose to normalize the gradient in each block:
\begin{eqnarray} \label{eq:grad_normalization}
\tg_{t, \mG_b} = g_{t, \mG_b} / \|g_{t, \mG_b}\|_2.
\end{eqnarray}
Then, we use $\tg_{t, \mG_b}$ to update first-order and second-order momentums $m_t$ and $v_t$, respectively. 
This technique was first introduced in \cite{yu2017block} for accelerating Adam in training deep neural networks. Using the per-block gradient normalization, the gradient clipping is no longer necessary. Ignoring the gradient magnitude makes the gradient descent methods more robust to vanishing and exploding gradients.

\subsection{Incorporate Nesterov's Momentum into LAMB}

In order to incorporate Nesterov's momentum, we first rewrite the step 11 in Algorithm~\ref{alg:lamb} as
\begin{eqnarray} 
x_{t + 1, \mG_b} & = & x_{t, \mG_b} - \eta_t\phi(\|x_{t, \mG_b}\|_2)\left[\frac{\beta_1}{\|r_{t, \mG_b} + \lambda x_{t, \mG_b}\|}\left(\frac{m_{t-1, \mG_b}/(1 - \beta_1^t)}{\sqrt{\tv_{t, \mG_b}} + \epsilon} + \lambda x_{t, \mG_b}\right) \right. \label{eq:lamb_expand_1} \\
&& \left. + \frac{1 - \beta_1}{\|r_{t, \mG_b} + \lambda x_{t, \mG_b}\|}\left(\frac{g_{t, \mG_b}/(1 - \beta_1^t)}{\sqrt{\tv_{t, \mG_b}} + \epsilon} + \lambda x_{t, \mG_b}\right) \right]. \label{eq:lamb_expand_2}
\end{eqnarray}
To apply the same trick as in Nesterov's momentum, first we substitute $m_{t, \mG_b}$ for $m_{t - 1, \mG_b}$ in (\ref{eq:lamb_expand_1}), and then we modify the normalization factors to ensure unit $\ell_2$-norm for both (\ref{eq:lamb_expand_1}) and (\ref{eq:lamb_expand_2}), leading to the following new update rule:
\begin{eqnarray} 
x_{t + 1, \mG_b} & = & x_{t, \mG_b} - \eta_t\phi(\|x_{t, \mG_b}\|_2)\left[\frac{\beta_1}{\|r_{t, \mG_b} + \lambda x_{t, \mG_b}\|}\left(r_t + \lambda x_{t, \mG_b}\right) \right. \nonumber \\
&& \left. + \frac{1 - \beta_1}{\|a_{t, \mG_b} + \lambda x_{t, \mG_b}\|}\left(a_{t, \mG_b} + \lambda x_{t, \mG_b}\right) \right], \label{eq:lans_expand_1}
\end{eqnarray}
where $a_{t, \mG_b} = \frac{g_{t, \mG_b}}{\sqrt{\tv_{t, \mG_b}} + \epsilon}$. Note that we remove the factor $1/(1 - \beta_1^t)$ in (\ref{eq:lamb_expand_2}) for (\ref{eq:lans_expand_1}), as this factor leads to a bias towards $g_{t, \mG_b}$ when the normalization is modified and regularization parameter $\lambda > 0$. 
Interestingly, the resultant update is simply a convex combination between LAMB updates with and without first-order momentum.
Combining (\ref{eq:lans_expand_1}) with (\ref{eq:grad_normalization}), we obtain Algorithm~\ref{alg:lans}. 

\begin{algorithm}[t]
\caption{LANS}
\label{alg:lans}
\begin{algorithmic}[1]
    \STATE {\bfseries Input:} stepsize sequence $\eta_t$; $0 < \beta_1, \beta_2 < 1$; scaling function $\phi$; $\epsilon > 0$; regularization parameter $\lambda$.
    \STATE {\bfseries Initialize:} $x_1 \in \mathbb{R}^d$; $m_0, v_0 = 0$.
    \FOR{$t = 1, \ldots, T$}
        \STATE Compute mini-batch stochastic gradient $g_{t}$
        \FOR{$b=1, 2, \dots, B$}
        \STATE{$\tg_{t, \mG_b} = g_{t, \mG_b} / \|g_{t, \mG_b}\|_2$}
        \STATE{$m_{t, \mG_b} = \beta_1m_{t-1, \mG_b} + (1 - \beta_1)\tg_{t, \mG_b}$}
        \STATE{$v_{t, \mG_b} = \beta_2v_{t-1, \mG_b} + (1 - \beta_2)\tg_{t, \mG_b}^2$}
        \STATE{$\tm_{t, \mG_b} = m_{t, \mG_b} / (1 - \beta_1^t)$}
        \STATE{$\tv_{t, \mG_b} = v_{t, \mG_b} / (1 - \beta_2^t)$}
        \STATE{compute ratios $r_{t, \mG_b} = \frac{\tm_{t, \mG_b}}{\sqrt{\tv_{t, \mG_b}} + \epsilon}$ and $c_{t, \mG_b} = \frac{\tg_{t, \mG_b}}{\sqrt{\tv_{t, \mG_b}} + \epsilon}$}
        \STATE{$d_{t, \mG_b} = \phi(\|x_{t, \mG_b}\|_2)\left[\frac{\beta_1}{\|r_{t, \mG_b} + \lambda x_{t, \mG_b}\|}(r_{t, \mG_b} + \lambda x_{t, \mG_b}) + \frac{1 - \beta_1}{\|c_{t, \mG_b} + \lambda x_{t, \mG_b}\|}(c_{t, \mG_b} + \lambda x_{t, \mG_b})\right]$}
        \STATE{$x_{t + 1, \mG_b} = x_{t, \mG_b} - \eta_td_{t, \mG_b}$}
     \ENDFOR
    \ENDFOR
\end{algorithmic}
\end{algorithm}

\subsection{Learning Rate Scheduler for Large Mini-Batch}

For large mini-batch optimization, warmup is usually used at the start of the training \cite{goyal2017accurate}. Goyal {\em et al.\/} \cite{goyal2017accurate} proposed to use a linear warmup in the beginning and return to the original learning rate schedule afterwards. For BERT pretraining, LAMB uses a learning rate schedule of form \cite{you2020reducing}
\begin{eqnarray} \label{eq:linear_warmup}
\eta_t = \begin{dcases}
    \eta\frac{t}{T_{warmup}},& \text{if } t \leq T_{warmup} \\
    \eta\frac{T - t}{T - T_{warmup}},              & \text{otherwise,}
\end{dcases}
\end{eqnarray}
where $\eta > 0$ is the maximum learning rate that the optimization algorithms use throughout the training and $T_{warmup}$ denotes the number of iterations in $warmup$ stage. 
It can be seen that $\eta_t$ gradually increases to $\eta$ when $t$ approaches $T_{warmup}$ and decreases to $0$ when $t \rightarrow T$.
In \cite{you2020reducing}, a square root scheduling rule is proposed for increasing mini-batch size: $\eta = \sqrt{k}\tilde{\eta}$, where $k$ is the mini-batch size and $\tilde{\eta}$ is a reference learning rate for a small mini-batch size. To achieve speedup using $\tau$ times larger mini-batch size, the number of training iterations $T$ is reduced by $\tau$ times while the learning rate $\eta$ is increased by $\sqrt{\tau}$ times. Using such scheduler, LAMB successfully scaled BERT pretraining up to a mini-batch size of 32K without any accuracy loss. 
For a larger mini-batch size, this square root scheduling breaks as it exceeds a maximum rate that does not depend on the mini-batch size. Thus, a smaller learning rate is used for a mini-batch size of 64K with a small degradation of accuracy. 

Considering that the learning rate is theoretically upper bounded by the inverse of the Lipschitz constant $L$ \cite{nesterov2004introductory,ghadimi2013stochastic,hardt2016train,you2020reducing} up to some small constants (e.g., $1$ or $2$), one cannot scale the learning rate indefinitely. We propose to add a $constant$ transient phase after the $warmup$ stage as shown below
\begin{eqnarray} \label{eq:linear_const_warmup}
\eta_t = \begin{dcases}
    \eta\frac{t}{T_{warmup}},& \text{if } t \leq T_{warmup} \\
    \eta,& \text{if } T_{warmup} < t \leq T_{warmup} + T_{const} \\
    \eta\frac{T - t}{T - T_{warmup} - T_{const}},              & \text{otherwise,}
\end{dcases}
\end{eqnarray}
where $T_{const}$ is the number of iterations in which a constant learning rate is used. This scheme allows the training to have sufficient progress even one cannot further increase $\eta$.
\begin{figure}[ht]
\begin{center}
\includegraphics[width=0.8\columnwidth]{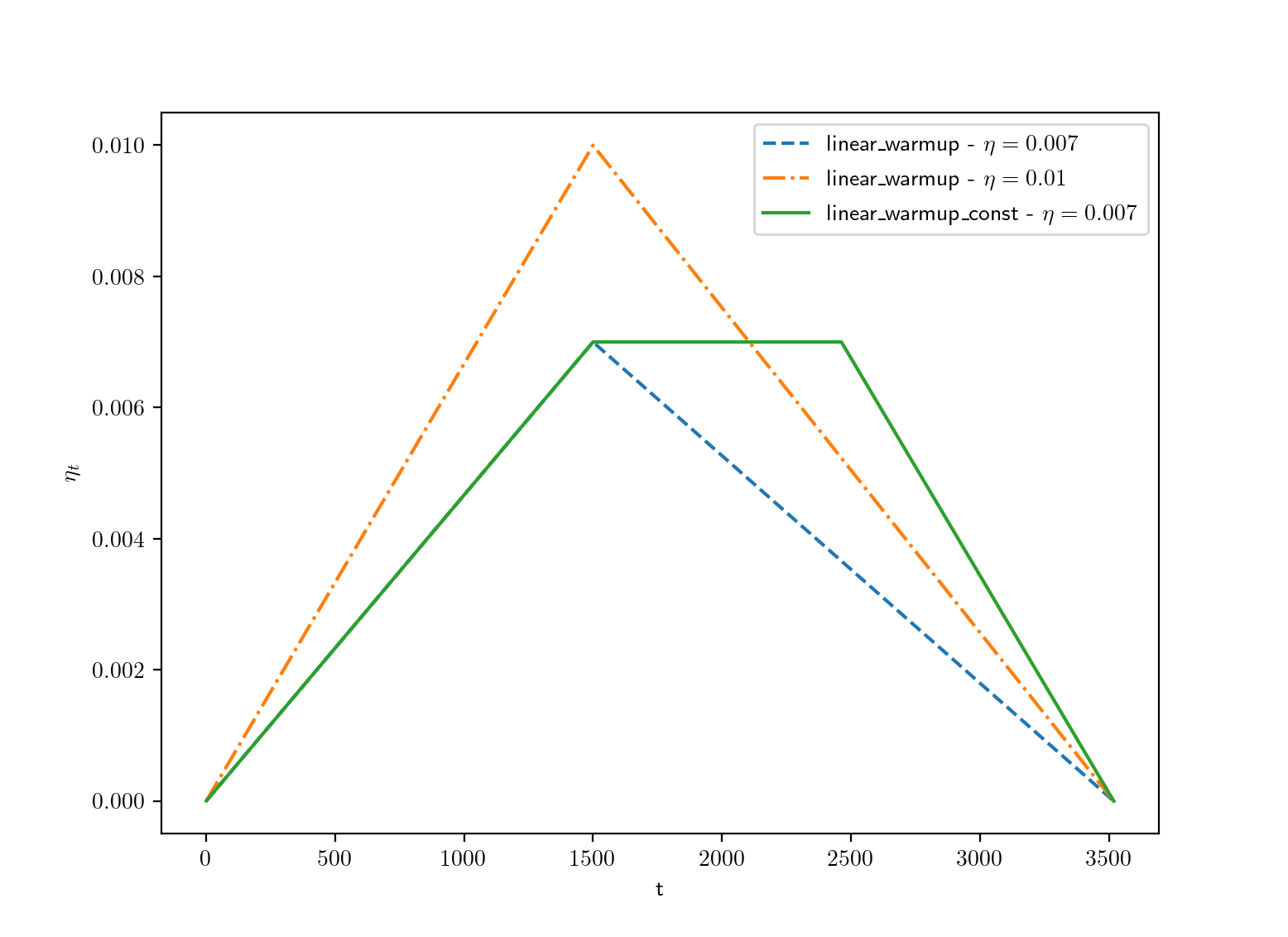} 
\caption{Visual illustrations of (\ref{eq:linear_warmup}) with $\eta=0.007, 0.01$ and (\ref{eq:linear_const_warmup}) with $\eta=0.007$. $T = 3519$, $T_{warmup}=1500$, and $T_{const} = 963$.
}
\label{fig:lr}
\end{center}
\end{figure}
Figure~\ref{fig:lr} shows visualization of (\ref{eq:linear_warmup}) with $\eta=0.007, 0.01$ and (\ref{eq:linear_const_warmup}) with $\eta=0.007$. $\eta=0.01$ refers to the ideal learning rate that we scaled the mini-batch size from 32K to 128K. However, $0.01$ has exceeded the maximum learning rate and results in divergence. Therefore, we have to use a smaller one such as $0.007$. Nonetheless, this smaller learning rate downgrades performance. In particular, the difference between areas under curve of (\ref{eq:linear_warmup}) with $\eta=0.007, 0.01$ is $5.28$. Using the proposed schedule (\ref{eq:linear_const_warmup}), we can reduce the difference to $1.91$.

\subsection{Data Sharding in Distributed Training}

In large-scale mini-batch training, the quality of the mini-batch plays an important role. In order to use large learning rate, one need to have as small gradient variance as possible. For example, random sampling with replacement results in a variance bound of $O(\frac{\sigma^2}{k})$ \cite{dekel2012optimal} while random sampling without replacement gives a better bound of $O(\frac{n-k}{k(n-1)}\sigma^2)$ \cite{li2014efficient}, where $\sigma^2$ is the upper bound of the gradient variance. It can be seen that the variance only goes to zero when $k \rightarrow \infty$ for random sampling with replacement while the variance is zero when $k=n$ for random sampling without replacement. Thus, random sampling without replacement results in better efficiency of using the same mini-batch size. In distributed training, to make sure that the mini-batch does not have redundant samples, we only grant each worker access to a shard of the dataset. Within each shard, random shuffling is used to construct the mini-batch samples. 

\section{Experiments}

In the experiment, we train a BERT-Large model on Wikipedia and BooksCorpus datasets. The experiment is conducted on 192 Amazon EC2 P3dn.24xlarge instances. There are $1536$ NVIDIA V100 GPUs in total. The preprocessed dataset is partitioned into $1536$ shards. The elastic fabric adapter (EFA) is enabled to improve the communication efficiency. 
We use LANS with the proposed learning rate schedule (\ref{eq:linear_const_warmup}). The training is divided into 2 stages: the first $3519$ iterations is trained with a short sequence length of $128$ and the last $782$ steps is trained with a longer sequence length of $512$. We use a mini-batch size of 96K and 33K for phases 1 and 2, respectively. 

Let $ratio_{warmup} = T_{warmup}/T_{stage_i}*100\%$ and $ratio_{const}=T_{const}/T_{stage_i}*100\%$ for $i$-th training stage. 
We use $ratio_{warmup} = 1.5 * ratio_{warmup_{64K}}$, where $ratio_{warmup_{64K}}$ is the warmup ratio used for LAMB with mini-batch sizes 64K/32K, and we select $ratio_{const}$ such that $ratio_{warmup} +  ratio_{const}=70\%$ and $ratio_{warmup} + ratio_{const}=30\%$ for stages 1 and 2 training, respectively. 
The hyper-parameters used in the experiments are shown in Table~\ref{tab:bert-pretrain-hyper-tbl}. 
\begin{table}[ht]
\begin{center}
\begin{tabular}{c|c|c|c}
 \hline 
 & $\eta$ & $ratio_{warmup}$ & $ratio_{const}$    \\\hline 
stage 1 & 0.00675 & 42.65\% & 27.35\% \\
stage 2 & 0.005 &19.2\% & 10.8\% \\\hline
\end{tabular}
\end{center}
\caption{Hyper-parameters used in LANS with mini-batch sizes 96K/33K.}
\label{tab:bert-pretrain-hyper-tbl}
\end{table} 
We can use larger learning rates such as 0.00725 in the first stage, but we observed that $\eta=0.00675$ gives better performance. For finetuning, we use AdamW optimizer \cite{loshchilov2017decoupled} with per-block gradient normalization (\ref{eq:grad_normalization}).
The experiment result in shown in Table~\ref{tab:bert-pretrain-tbl}. 
\begin{table}[ht]
\begin{center}
\begin{tabular}{c|c|c|c|c|c}
 \hline 
 & batch size & steps & F1 score on dev set & TPUs/GPUs & time   \\\hline 
LAMB \cite{you2020reducing} & 64K/32K & 8599 & 90.58 & 1024 TPUs & 76.2m \\ 
LAMB \cite{you2020reducing} & 96K/33K & 4301  & diverge & 1536 GPUs &  N/A \\ 
LANS & 96K/33K & 4301 & 90.60 & 1536 GPUs & 53.6m \\ \hline 
\end{tabular}
\end{center}
\caption{Experiment results on BERT pretraining. The F1 score on SQuAD-v1.1 development set is used as the evaluation metric. The result of LANS is compared to the one of LAMB from Table~1 in \cite{you2020reducing}. }
\label{tab:bert-pretrain-tbl}
\end{table} 
As can be seen, the proposed methods only need $4301$ iterations and finish the BERT pretraining in 53.6 minutes, while LAMB fails to further scale up the mini-batch size in BERT training.   
On the other hand, when the model is trained with $4301$ steps, the square root scheduling rule suggests larger mini-batch sizes of 128K and 64K for stages 1 and 2, respectively. 
With the proposed methods, we are able to achieve the target accuracy using much smaller mini-batch sizes. This further reduces the total computational workload and training time. 

\section{Conclusion}
In this paper, we propose an accelerated large batch method called LANS. LANS employs block-wise gradient normalization and Nesterov's momentum. By identifying the insufficiency the of linear warmup learning rate schedule for large mini-batch training, we introduce a new learning rate scheduler that adopts a constant learning rate for few epochs after the warmup phase. The empirical evaluation shows that the proposed methods scale the BERT pretraining to mini-batch sizes of 96K and 33K for first and second training stages, respectively. And, they only use 54 minutes to complete the BERT training on 192 Amazon EC2 P3dn.24xlarge instances.

\bibliography{lans}
\bibliographystyle{plain}

\end{document}